\pdfoutput=1

\documentclass[11pt]{article}

\usepackage{ACL2023}

\usepackage{times}
\usepackage{latexsym}

\usepackage[T1]{fontenc}

\usepackage[utf8]{inputenc}

\usepackage{microtype}

\usepackage{inconsolata}

\usepackage{subfigure}
\usepackage{xfrac}
\usepackage{colortbl}
\usepackage{booktabs}  
\usepackage{amsfonts}  
\usepackage{nicefrac}
\usepackage{latexsym}
\usepackage{multirow}
\usepackage{amsmath}
\usepackage{tabularx}
\usepackage{makecell}
\usepackage{xcolor}
\definecolor{gred}{RGB}{219,68,55}
\definecolor{gblue}{RGB}{66,133,244}
\definecolor{gyellow}{RGB}{244,180,0}
\definecolor{ggreen}{RGB}{85,157,88}
\definecolor{ggrey}{RGB}{115,115,115}
\definecolor{na}{gray}{0.9}

\newcommand{\colorR}[1]{\textcolor{gred}{#1}}
\newcommand{\colorG}[1]{\textcolor{ggreen}{#1}}
\newcommand{\colorB}[1]{\textcolor{gblue}{#1}}
\newcommand{\colorY}[1]{\textcolor{gyellow}{#1}}

\newcommand*{\email}[1]{\texttt{#1}}

\title{ReactIE: Enhancing Chemical Reaction Extraction with Weak Supervision}

\author{
\bf Ming Zhong 
\quad Siru Ouyang
\quad Minhao Jiang \\
\bf \quad Vivian Hu
\quad Yizhu Jiao
\quad Xuan Wang
\quad Jiawei Han
\\
University of Illinois Urbana-Champaign, IL, USA \\
\email{\{mingz5, siruo2, minhaoj2, vivianhu2, yizhuj2, xwang174, hanj\}}@illinois.edu}

\begin{document}
\maketitle
\begin{abstract}
Structured chemical reaction information plays a vital role for chemists engaged in laboratory work and advanced endeavors such as computer-aided drug design. Despite the importance of extracting structured reactions from scientific literature, data annotation for this purpose is cost-prohibitive due to the significant labor required from domain experts. Consequently, the scarcity of sufficient training data poses an obstacle to the progress of related models in this domain. In this paper, we propose \textsc{ReactIE}, which combines two weakly supervised approaches for pre-training. Our method utilizes frequent patterns within the text as \textit{linguistic cues} to identify specific characteristics of chemical reactions. Additionally, we adopt synthetic data from patent records as distant supervision to incorporate \textit{domain knowledge} into the model. Experiments demonstrate that \textsc{ReactIE} achieves substantial improvements and outperforms all existing baselines.
\end{abstract}

\section{Introduction}

The integration of advanced Natural Language Processing (NLP) techniques in the field of chemistry has been gaining significant attention in both academia and industry \cite{DBLP:conf/bcb/WangGWSH19,fabian2020molecular, DBLP:journals/corr/abs-2010-09885}.
By formulating applications in chemistry as molecular representation \cite{DBLP:conf/mlhc/ShinPKH19, DBLP:conf/iclr/WangLJCJ0B22}, information extraction \cite{vaucher2020automated, DBLP:conf/emnlp/WangHSGX021, DBLP:conf/bibm/WangHJZXLJBH22}, and text generation \cite{DBLP:journals/corr/abs-2204-11817} tasks, NLP approaches provide new avenues for effective understanding and analysis of chemical information.
In particular, we focus on the chemical reaction extraction task, as it can serve as a valuable reference for chemists to conduct bench experiments \cite{DBLP:journals/jcisd/GuoIGQCJB22}.

Despite the abundance of text describing chemical reactions in the scientific literature, the conversion to a structured format remains a major challenge.
One approach is the utilization of domain experts to manually extract chemical reactions, resulting in several commercial reaction databases, such as Reaxys \cite{DBLP:journals/jcisd/Goodman09} and SciFinder \cite{gabrielson2018scifinder}.
However, this method is associated with significant time and labor costs, as well as the issue of restricted access to these resources.

\begin{figure}[t]
    \centering
    \includegraphics[width=1\linewidth]{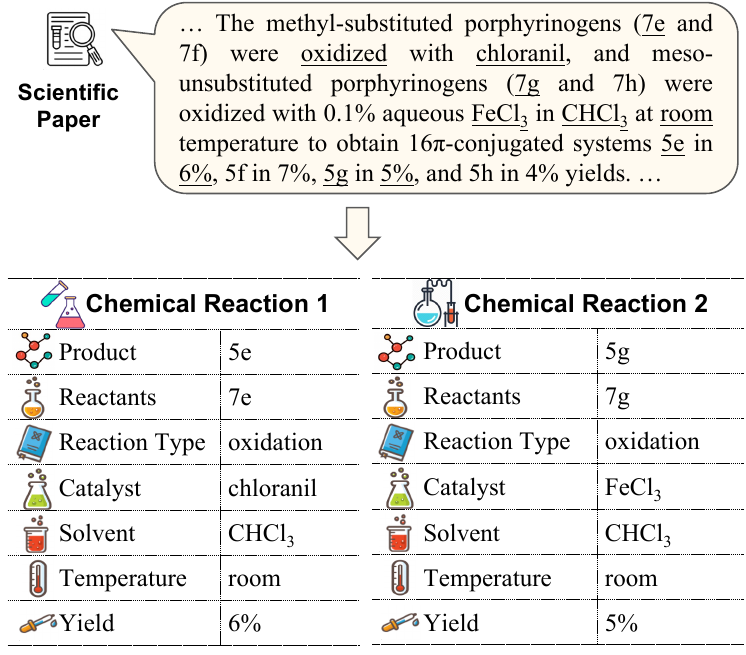}
    \caption{An example of the chemical reaction extraction task.
    This figure depicts two out of the four chemical reactions present in the text for simplicity. The passage is drawn from \citet{ahmad2015synthesis}.}
    \label{fig:task}
    \vspace{-2.2mm}
\end{figure}

Subsequently, research efforts concentrated on automated systems, including \textsc{Opsin} \cite{DBLP:phd/ethos/Lowe12} and \textsc{ChemRxnBERT} \cite{DBLP:journals/jcisd/GuoIGQCJB22}.
\textsc{Opsin} is a heuristic-based system that employs a complex set of rules to identify the reaction roles.
While it is effective for well-formatted text, \textsc{Opsin}'s performance is limited in scientific literature due to its sensitivity to variations in language use.
In contrast, \citet{DBLP:journals/jcisd/GuoIGQCJB22} obtained \textsc{ChemRxnBERT} by pre-training with language modeling on chemistry journals, however, the model performance is constrained by the small size of the training set during fine-tuning.
This raises the question of how to effectively utilize large-scale unlabeled data for this task, which remains an under-explored area.

In this paper, we present \textsc{ReactIE}, a pre-trained model for chemical reaction extraction.
In light of the clear gap between prevalent pre-training tasks and the applications in the field of chemistry, we propose two weakly supervised methods to construct synthetic data for pre-training.
Intuitively, humans can infer certain roles in chemical reactions from \textbf{linguistic cues}. As shown in Figure \ref{fig:task}, we can identify ``\textit{5e}'' as the product from the semantic meaning of the phrase ``\textit{to obtain 5e}''.
To this end, we mine frequent patterns from texts as linguistic cues and inject them into the model.
Furthermore, \textbf{domain knowledge} also plays a crucial role in this task. For example, the accurate identification of ``\textit{chloranil}'' as a catalyst rather than a reactant in Figure \ref{fig:task} requires a deep understanding of related compounds.
To address this, we incorporate domain knowledge into \textsc{ReactIE} by utilizing patent literature as distant supervision.
By pre-training on these acquired synthetic data, \textsc{ReactIE} maintains consistency with downstream objectives.

Experimentally, \textsc{ReactIE} achieves state-of-the-art performance, improving F$_1$ scores by 14.9 and 2.9 on the two subtasks, respectively.
Moreover, we conduct ablation studies to examine the contributions of the proposed methods.
Fine-grained analyses are performed to investigate the effects of pre-training strategies on different reaction roles.
Our findings suggest that linguistic cues are crucial for extracting products and numbers, while chemical knowledge plays an essential role in understanding catalysts, reactants, and reaction types.

\section{Preliminary}

\subsection{Task Formulation}
Given a text $D$, the goal of this task is to extract all the structured chemical reactions $\mathcal{S}$ in $D$, where each $S \in \mathcal{S}$ contains $n$ role-argument pairs \{($r_1$, $a_1$), $\cdots$, ($r_n$, $a_n$)\}. The roles are 8 pre-defined attributes in a chemical reaction, including \textit{product}, \textit{reactant}, \textit{catalyst}, \textit{solvent}, \textit{reaction type}, \textit{temperature}, and \textit{yield}. Each $S$  does not include the roles that are not present in the original text. Definitions for each role are included in Appendix \ref{sec:scheme}.

\subsection{Workflow for IE System}

From the perspective of the model, existing systems typically follow a two-step pipeline:

1) \textbf{ Product Extraction}: In chemical reactions, the product is the central factor as the same reactants can yield varying products depending on the reaction conditions.
Therefore, the IE systems first extract all the products in $D$ to determine the number of chemical reactions, i.e., the number of $S$.
This step can also be used to extract passages in a scientific paper that contain chemical reactions.

2) \textbf{ Role Extraction}: Given the original text $D$ and the specific product, the IE systems are required to capture the relationship between the entities in $D$ and the product, extract the corresponding reaction roles, and output the final $\mathcal{S}$.

\begin{figure*}[t]
    \centering
    \includegraphics[width=0.866\linewidth]{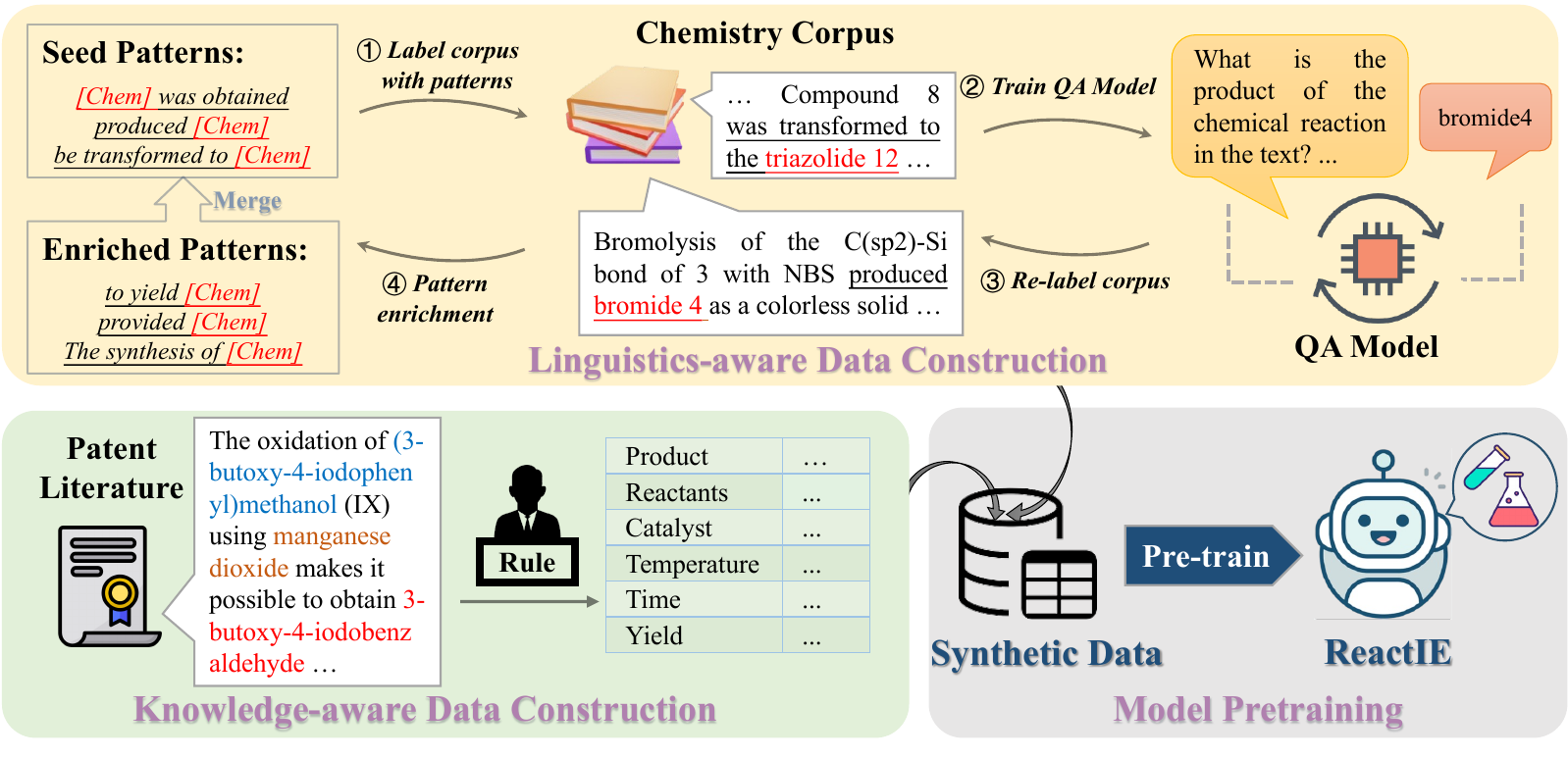}
    \caption{Overview of \textsc{ReactIE}. We propose linguistics-aware and knowledge-aware methods to construct synthetic data, thus bridging the gap between the objectives of pre-training and the chemical reaction extraction task.}
    \label{fig:model}
    \vspace{-2.9mm}
\end{figure*}

\section{\textsc{ReactIE} Framework}

\subsection{Reformulation} Previous studies have defined this task as a sequence labeling problem\footnote{The reaction roles are captured using ``BIO'' scheme.}. However, this approach could be inadequate in certain cases. For instance, the final argument may be an alias, abbreviation, or pronoun of a compound in $D$, or the necessary conversion of words should be made (as illustrated in Figure \ref{fig:task}, ``\textit{oxidized}''  $\rightarrow$ ``\textit{oxidation}'').

In light of these limitations, we reformulate the chemical reaction extraction task as a Question Answering (QA) problem, utilizing the pre-trained generation model FLAN-T5 \cite{DBLP:journals/corr/abs-2210-11416} as the backbone.
For product extraction, the input question is ``\textit{What are the products of the chemical reactions in the text?}''.
For role extraction, such as catalyst, the corresponding question is ``\textit{If the final product is X, what is the catalyst for this chemical reaction?}''.
In this unified QA format, we present the pre-training stage of \textsc{ReactIE} as follows.

\subsection{Pre-training for \textsc{ReactIE}}

Given the clear discrepancy between prevalent pre-training tasks such as language modeling and the task of chemical reaction extraction, we propose two weakly supervised methods for constructing synthetic data to bridge this gap.

\paragraph{Linguistics-aware Data Construction}
Intuitively, it is possible for humans to infer certain properties of a chemical reaction, even without any prior knowledge of chemistry.
As an example, consider the sentence ``\textit{Treatment of 13 with lithium benzyl oxide in THF afforded the dihydroxybenzyl ester 15}'' \cite{dushin1992total}. We can identify that ``\textit{13}'' and ``\textit{lithium benzyl}'' are the reactants, and ``\textit{dihydroxybenzyl ester 15}'' is the end product, without knowing any specific compounds involved.
This can be achieved by utilizing \textbf{linguistic cues} such as the semantics of phrases and the structure of sentences to extract the arguments.

Inspired by this, we leverage frequent patterns \cite{DBLP:conf/kdd/JiangSCRKH017} in the text that describes specific reaction roles as linguistic cues.
Take product extraction as an example, we first replace the chemical with a special token ``[Chem]'' using \textsc{ChemDataExtractor} \cite{DBLP:journals/jcisd/SwainC16}, and then manually create a set of seed patterns, such as \textit{the produced \colorR{[Chem]}}, \textit{conversion of [Chem] to \colorR{[Chem]}}, etc.
The red \colorR{[Chem]} indicates that the chemical here is the product of a reaction.
As shown in Figure \ref{fig:model}, based on seed patterns and a  chemistry corpus, we construct synthetic data as:

\noindent 1) Seed patterns are used to annotate the chemical corpus, resulting in training data containing labels.

\noindent 2) Continue training Flan-T5 in QA format on the data from the previous step.

\noindent 3) Use the QA model to re-label the entire corpus.

\noindent 4) The most frequent patterns are mined from the data in step 3 as the enriched pattern set.

By merging  the seed patterns in the first step with the enriched patterns, we can iteratively repeat the process and collect reliable data containing multiple linguistic cues. More examples and details can be found in Appendix \ref{sec:pattern} and Table \ref{tab:pattern}.

\paragraph{Knowledge-aware Data Construction}
In addition to utilizing linguistic cues, a deep understanding of chemical reactions and terminology is  imperative for accurately extracting information from texts.
This is exemplified in the case presented in Figure \ref{fig:task}, in which the roles of compounds such as ``\textit{chloranil}'', ``\textit{FeCl$_3$}'' and ``\textit{CHCl$_3$}'' as reactants, catalysts, or solvents cannot be inferred without prior knowledge.
In light of this, we propose the integration of \textbf{domain knowledge} into \textsc{ReactIE} through the synthetic data derived from patent records.

The text within patent documents is typically well-formatted, allowing for the extraction of structured chemical reactions through the well-designed rules incorporating multiple chemical principles and associated knowledge bases \cite{DBLP:phd/ethos/Lowe12}.
To utilize this, we adopt datasets extracted from the U.S. patent literature by OPSIN \cite{lowe2018chemical} as our synthetic data.
We focus on 4 reaction roles (product, reactant, catalyst, and solvent) that are most relevant to chemistry knowledge.

\paragraph{Training Paradigm}
The methods outlined above enable the acquisition of a substantial amount of synthetic data.
We then proceed to conduct pre-training by building upon the FLAN-T5 model in a text-to-text format. The input contains questions $q_i$ specific to a reaction role $r_i$ and text $D$, and the output is the corresponding argument $a_i$ or ``None''.
After pre-training, the unsupervised version of \textsc{ReactIE} acquires the capability to extract structured chemical reactions. To further improve it, we also perform fine-tuning on an annotated dataset to attain a supervised version of \textsc{ReactIE}.

\section{Experiments}

\subsection{Experimental Setup}

\paragraph{Datasets} We use Reaction Corpus  \cite{DBLP:journals/jcisd/GuoIGQCJB22} which includes 599/96/111 annotated chemical reactions in training, dev, and test sets.
The input is a paragraph in scientific papers and the output consists of multiple structured chemical reactions in the text.
This corpus is designed to evaluate two subtasks, product extraction, and role extraction.

\paragraph{Baselines}
We compare the performance of \textsc{ReactIE} with several state-of-the-art baselines, including \textsc{Opsin}, \textsc{BiLSTM-CRF} \cite{DBLP:journals/corr/HuangXY15}, \textsc{Bert} \cite{DBLP:conf/naacl/DevlinCLT19}, \textsc{BioBert} \cite{DBLP:journals/bioinformatics/LeeYKKKSK20}, \textsc{ChemBert}, and \textsc{ChemRxnBert} \cite{DBLP:journals/jcisd/GuoIGQCJB22}. \textsc{Opsin} is an unsupervised rule-based system while the variants of BERT are pre-trained on different domain-specific corpora.

\paragraph{Implementation Details}
We use ``google/flan-t5-large” as the backbone model in all experiments.
For linguistics-aware data construction, we perform 3 iterations on 18,894 chemical journals and end up with 92,371 paragraphs containing the linguistic cues of product, temperature, yield, and time.
Other reaction roles are excluded because they do not have sufficient patterns to ensure the reliability of the data.
For knowledge-aware data construction, excessively long (> 256 words) and short (< 8 words) texts, as well as samples where the arguments do not appear in the original text, are filtered to yield 100,000 data.
We train \textsc{ReactIE} for 1 epoch with 0.1 label smoothing on a total of 192,371 samples.
For both pre-training and fine-tuning, we set the batch size to 16 with 5e-5 as the learning rate.
All results are the performance of the checkpoints selected by the dev set.

\renewcommand\arraystretch{1}
\begin{table}
\centering \footnotesize
\tabcolsep0.12 in
\begin{tabular}{lccc}
\toprule
\textbf{Models} & \textbf{P}~(\%) & \textbf{R}~(\%) & \textbf{F}~(\%) \\  
\midrule

\multicolumn{4}{c}{\textit{Unsupervised}} \\

\midrule

\textsc{Opsin} & 18.8 & 5.4 & 8.4 \\
\cellcolor[rgb]{ .851,  .851,  .851}\textsc{ReactIE} & \cellcolor[rgb]{ .851,  .851,  .851}\textbf{69.7} & \cellcolor[rgb]{ .851,  .851,  .851}\textbf{53.5} & \cellcolor[rgb]{ .851,  .851,  .851}\textbf{60.5} \\

\midrule

\multicolumn{4}{c}{\textit{Supervised}} \\

\midrule

\textsc{BiLSTM} & 52.4 & 46.7 & 49.4 \\
\textsc{BiLSTM} (w/ CRF) & 54.3 & 49.1 & 51.6 \\
\textsc{BERT} & 78.8 & 56.8 & 66.0 \\
\textsc{BioBERT} & 76.4 & 61.3 & 68.0 \\
\textsc{ChemBERT} & 84.6 & 69.4 & 76.2 \\

\cellcolor[rgb]{ .851,  .851,  .851}\textsc{FlanT5} & \cellcolor[rgb]{ .851,  .851,  .851}88.0 & \cellcolor[rgb]{ .851,  .851,  .851}83.2 & \cellcolor[rgb]{ .851,  .851,  .851}85.5 \\

\cellcolor[rgb]{ .851,  .851,  .851}\textsc{ReactIE} & \cellcolor[rgb]{ .851,  .851,  .851}\textbf{94.2} & \cellcolor[rgb]{ .851,  .851,  .851}\textbf{88.2} & \cellcolor[rgb]{ .851,  .851,  .851}\textbf{91.1} \\
\cellcolor[rgb]{ .851,  .851,  .851}\quad - \textit{linguistics cues} & \cellcolor[rgb]{ .851,  .851,  .851}89.8 & \cellcolor[rgb]{ .851,  .851,  .851}84.7 & \cellcolor[rgb]{ .851,  .851,  .851}87.2 \\
\cellcolor[rgb]{ .851,  .851,  .851}\quad - \textit{domain knowledge} & \cellcolor[rgb]{ .851,  .851,  .851}92.6 & \cellcolor[rgb]{ .851,  .851,  .851}87.1 & \cellcolor[rgb]{ .851,  .851,  .851}89.8 \\

\bottomrule
\end{tabular}
\caption{Results for product extraction. The results presented in the gray background correspond to the performance of \textsc{ReactIE} and its ablation studies.}
\label{tab:product}
\vspace{-2mm}
\end{table}

\subsection{Experimental Results}

\paragraph{Results for Product Extraction}
The first part of Table \ref{tab:product} presents the results under the unsupervised setting.
\textsc{Opsin} performs poorly in the scientific paper domain due to its sensitivity to language usage.
In contrast, \textsc{ReactIE} demonstrates superior extraction capabilities after pre-training and outperforms the fully supervised BiLSTM (w/ CRF).

Under the supervised setting, \textsc{ReactIE} attains state-of-the-art performance with a significant margin, achieving a 14.9 increase in F$_1$ scores compared to \textsc{ChemBERT}.
While our backbone model, \textsc{FlanT5}, shows outstanding results, our proposed methods can lead to further gains (85.5 $\Rightarrow$ 91.1 F$_1$).
Ablation studies highlight the importance of linguistics-aware pre-training over in-domain knowledge in the product extraction subtask.
This finding also supports the advantages of pre-trained language models (\textsc{FlanT5}) over domain-specific models (\textsc{ChemBERT}), as the writers have provided sufficient linguistic cues for the products of chemical reactions when describing them.

\paragraph{Results for Role Extraction} As listed in Table \ref{tab:role}, \textsc{ReactIE} also beats the previous best model \textsc{ChemRxnBERT} by 2.9 F$_1$ score for the role extraction subtask.
In comparison to the product, the accurate extraction of other reaction roles from the original text necessitates a greater level of in-domain knowledge.
Specifically, the model performance decreases slightly (81.6 $\Rightarrow$ 80.6 F$_1$) when linguistics-aware pre-training is removed, and substantially by 4.4 (81.6 $\Rightarrow$ 77.2 F$_1$) when knowledge-aware pre-training is no longer incorporated.
The results of these two subtasks reveal that our proposed approaches are complementary and indispensable in enabling \textsc{ReactIE} to fully comprehend chemical reactions. 
Together, they contribute to a deeper understanding of the task from both linguistic and chemical knowledge perspectives.

\renewcommand\arraystretch{1}
\begin{table}
\centering \footnotesize
\tabcolsep0.12 in
\begin{tabular}{lccc}
\toprule
\textbf{Models} & \textbf{P}~(\%) & \textbf{R}~(\%) & \textbf{F}~(\%) \\

\midrule

\textsc{BERT} & 69.2 & 69.2 & 69.2 \\
\textsc{BioBERT} & 73.3 & 75.5 & 74.3 \\
\textsc{ChemBERT} & 77.0 & 76.4 & 76.7 \\
\textsc{ChemRxnBERT} & 79.3 & 78.1 & 78.7 \\

\cellcolor[rgb]{ .851,  .851,  .851}\textsc{FlanT5} & \cellcolor[rgb]{ .851,  .851,  .851}76.1 & \cellcolor[rgb]{ .851,  .851,  .851}75.4 & \cellcolor[rgb]{ .851,  .851,  .851}75.8 \\

\cellcolor[rgb]{ .851,  .851,  .851}\textsc{ReactIE} & \cellcolor[rgb]{ .851,  .851,  .851}\textbf{80.8} & \cellcolor[rgb]{ .851,  .851,  .851}82.5 & \cellcolor[rgb]{ .851,  .851,  .851}\textbf{81.6} \\
\cellcolor[rgb]{ .851,  .851,  .851}\quad - \textit{linguistics cues} & \cellcolor[rgb]{ .851,  .851,  .851}78.1 & \cellcolor[rgb]{ .851,  .851,  .851}\textbf{83.3} & \cellcolor[rgb]{ .851,  .851,  .851}80.6 \\

\cellcolor[rgb]{ .851,  .851,  .851}\quad - \textit{domain knowledge} & \cellcolor[rgb]{ .851,  .851,  .851}74.8 & \cellcolor[rgb]{ .851,  .851,  .851}79.8 & \cellcolor[rgb]{ .851,  .851,  .851}77.2 \\

\bottomrule
\end{tabular}
\caption{Results for role extraction.}
\label{tab:role}
\end{table}

\begin{figure}[t]
    \centering
    \includegraphics[width=0.9\linewidth]{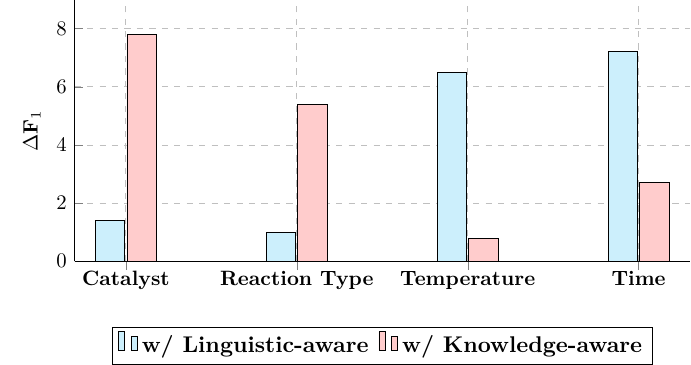}
    \caption{The impact of two pre-training strategies on different chemical reaction roles. The Y-axis shows the F$_1$ improvement compared to the backbone model.}
    \label{fig:analysis}
    \vspace{-2mm}
\end{figure}

\paragraph{Analysis for Reaction Roles}

To further investigate the effect of our  pre-training strategies, we present $\Delta$F$_1$ scores on different reaction roles after equipping the two methods separately in Figure \ref{fig:analysis}.
We can observe that these two strategies assist the model by concentrating on distinct aspects of chemical reactions.
Linguistic-aware pre-training primarily improves performance in reaction roles related to numbers, as these numbers tend to appear in fixed meta-patterns.
In contrast, knowledge-related pre-training significantly enhances the results of catalyst and reaction type, which require a chemical background for accurate identification.
Overall, the combination of both approaches  contributes to the exceptional performance of \textsc{ReactIE} in the chemical reaction extraction task.

\section{Conclusion}

In this paper, we present \textsc{ReactIE}, an automatic framework for extracting chemical reactions from the scientific literature. 
Our approach incorporates linguistic and chemical knowledge into the pre-training.
Experiments show that \textsc{ReactIE} achieves state-of-the-art results by a large margin.

\section*{Limitations}

We state the limitations of this paper from the following three aspects:

1) Regarding linguistics-aware data construction, we only perform seed-guided pattern enrichment for four reaction roles (product, yield, temperature, and time, see Table \ref{tab:pattern}) due to the lack of sufficient reliable patterns for other roles. Incorporating more advanced pattern mining methods \cite{DBLP:conf/kdd/Li0ZQHG018, DBLP:conf/emnlp/0001LDWZ0022} may alleviate this issue and discover more reliable linguistic cues, which we leave for future work.

2) As in the previous work, we adopt a fixed reaction scheme to extract structured chemical reaction information.
However, there are always new informative roles in the text \cite{DBLP:journals/corr/abs-2211-01577}, such as experimental procedures \cite{vaucher2021inferring}, so how to predict both roles and arguments without being limited to a fixed scheme could be a meaningful research topic.

3) \textsc{ReactIE} is capable of detecting chemical reactions within scientific literature by predicting if a given passage contains a product.
However, accurate text segmentation of a paper remains an unresolved and crucial issue.
Incomplete segmentation may result in the failure to fully extract reaction roles, while excessively long segmentation may negatively impact the model performance.
Therefore, integrating a text segmentation module into the existing two-step pipeline may be the next stage in the chemical reaction extraction task.

\section*{Acknowledgements}
We thank anonymous reviewers for their valuable comments and suggestions. Research was supported in part by US DARPA KAIROS Program No. FA8750-19-2-1004 and INCAS Program No. HR001121C0165, National Science Foundation IIS-19-56151, IIS-17-41317, and IIS 17-04532, and the Molecule Maker Lab Institute: An AI Research Institutes program supported by NSF under Award No. 2019897, and the Institute for Geospatial Understanding through an Integrative Discovery Environment (I-GUIDE) by NSF under Award No. 2118329. Any opinions, findings, and conclusions or recommendations expressed herein are those of the authors and do not necessarily represent the views, either expressed or implied, of DARPA or the U.S. Government. The views and conclusions contained in this paper are those of the authors and should not be interpreted as representing any funding agencies.

\bibliography{anthology,custom}
\bibliographystyle{acl_natbib}

\appendix

\clearpage

\section{Reaction Scheme}
\label{sec:scheme}

We adopt the same reaction scheme as in the previous study, including 8 pre-defined reaction roles to cover the source chemicals, the outcome, and the  conditions of a chemical reaction. To help better understand each reaction role, we include the detailed descriptions of the reaction scheme in \citet{DBLP:journals/jcisd/GuoIGQCJB22} as a reference in Table \ref{tab:scheme}.

\section{Pattern Enrichment in Linguistics-aware Data Construction}
\label{sec:pattern}

Table \ref{tab:pattern} provides examples of seed and enriched patterns for the product, yield, temperature, and time.
In each iteration, we extract n-grams ($n$ = \{$2, \cdots, 6$\}) containing the product (\colorR{[Chem]}), yield (\colorB{[Num]}), temperature (\colorG{[Num]}), and time (\colorY{[Num]}) from the corpus re-labeled by the QA model and remove the redundant patterns. We manually review and select reliable patterns and merge them into the pattern set of the previous iteration.

\renewcommand\arraystretch{1.1}
\begin{table}
\centering \footnotesize
\tabcolsep0.08 in
\begin{tabular}{ll}
\toprule
\textbf{Reaction Role} & \multicolumn{1}{c}{\textbf{Description}} \\

\midrule

\textbf{Product} & \multicolumn{1}{m{5cm}}{Chemical substance that is the final outcome (major product) of the reaction} \\

\midrule

\textbf{Reactants} & \multicolumn{1}{m{5cm}}{Chemical substances that contribute heavy atoms to the product} \\

\midrule

\textbf{Catalyst} & \multicolumn{1}{m{5cm}}{Chemical substances that participate in the reaction but do not contribute heavy atoms (e.g., acid, base, metal complexes)} \\

\midrule

\textbf{Solvent} & \multicolumn{1}{m{5cm}}{Chemical substances that are used to dissolve/mix other chemicals, typically quantified by volume and used in superstoichiometric amounts (e.g., water, toluene, THF)} \\

\midrule

\textbf{Temperature} & \multicolumn{1}{m{5cm}}{Temperature at which the reaction occurs} \\

\midrule

\textbf{Time} & \multicolumn{1}{m{5cm}}{Duration of the reaction performed} \\

\midrule

\textbf{Reaction Type} & \multicolumn{1}{m{5cm}}{Descriptions about the type of chemical reaction} \\

\midrule

\textbf{Yield} & \multicolumn{1}{m{5cm}}{ Yield of the product} \\

\bottomrule
\end{tabular}
\caption{Reaction scheme used in this paper.}
\label{tab:scheme}
\end{table}

\renewcommand\arraystretch{1.1}
\begin{table*}
\centering \footnotesize
\tabcolsep0.2 in
\begin{tabular}{ll}
\toprule
\textbf{Seed Patterns} (completed set) & \textbf{Enriched Patterns} (randomly sampled set)  \\  
\midrule
\multicolumn{2}{c}{\textit{Product}} \\
\midrule
produced \colorR{[Chem]} & to yield \colorR{[Chem]} \\ 

\colorR{[Chem]} be obtained & provided \colorR{[Chem]} \\

[Chem] be transformed to \colorR{[Chem]} & synthesis of \colorR{[Chem]} \\
\colorR{[Chem]} be systhesized from [Chem] & \colorR{[Chem]} be prepared from [Chem] \\
conversion of [Chem] to \colorR{[Chem]} & desired \colorR{[Chem]} \\

\midrule
\multicolumn{2}{c}{\textit{Yield}} \\
\midrule

in \colorB{[Num]} \% yield & at \colorB{[Num]} \% conversion \\
a yield of \colorB{[Num]} \% & in \colorB{[Num]} \% isolated yield \\
( \colorB{[Num]} \% yield ) & ( \colorB{[Num]} \% overall ) \\

\midrule
\multicolumn{2}{c}{\textit{Temperature}} \\
\midrule

at \colorG{[Num]} °C & ( \colorG{[Num]} °C ) \\
at \colorG{[Num]} K & a reaction temperature of \colorG{[Num]} °C \\
at \colorG{[Num]} OC & from \colorG{[Num]} to \colorG{[Num]} °C \\

\midrule
\multicolumn{2}{c}{\textit{Time}} \\
\midrule

for \colorY{[Num]} h & over \colorY{[Num]} h \\
for \colorY{[Num]} min & within \colorY{[Num]} h \\
for \colorY{[Num]} seconds & ( [Num] °C, \colorY{[Num]} h) \\
after \colorY{[Num]} h & for \colorY{[Num]} days \\

\bottomrule
\end{tabular}
\caption{Examples of Seed and enriched meta-patterns for the product, yield, and temperature. All seed patterns (3-5 patterns per role) are included here, and random samples from the enriched patterns are used as examples (20-50 patterns total per role). The red \colorR{[Chem]} indicates that the corresponding chemical is a product. The blue \colorB{[Num]} denotes that the number is a yield, green \colorG{[Num]} represents the temperature, and yellow \colorY{[Num]} is the reaction time.}
\label{tab:pattern}
\end{table*}

\end{document}